\documentclass{article}
\usepackage[utf8]{inputenc}
\usepackage{authblk}
\usepackage{setspace}
\usepackage[margin=1.25in]{geometry}
\usepackage{graphicx}
\graphicspath{ {./figures/} }
\usepackage{subcaption}
\usepackage{amsmath}

\usepackage[style=nejm, 
citestyle=numeric-comp,
sorting=none]{biblatex}
\title{HapMorph: A Pneumatic Framework for Multi-Dimensional Haptic Property Rendering}

\author[1*]{Rui~Chen}
\author[1]{Domenico~Chiaradia}
\author[1]{Antonio~Frisoli}
\author[1]{Daniele~Leonardis}

\affil[1]{Institute of Mechanical Intelligence, School of Advanced Studies Sant'Anna (SSSA), 56127 Pisa, Italy.}
\affil[*]{Corresponding author. Email: rui.chen@santannapisa.it}

\date{}

\begin{document}

\maketitle

\begin{abstract}
Haptic interfaces that can simultaneously modulate multiple physical properties remain a fundamental challenge in human-robot interaction. Existing systems typically allow the rendering of either geometric features or mechanical properties, but rarely both, within wearable form factors. Here, we introduce HapMorph, a pneumatic framework that enables continuous, simultaneous modulation of object size and stiffness through antagonistic fabric-based pneumatic actuators (AFPAs). We implemented a HapMorph protoytpe designed for hands interaction achieving size variation from 50 to 104~mm, stiffness modulation up to 4.7~N/mm and mass of the wearable parts of just 21~g. Through systematic characterization, we demonstrate decoupled control of size and stiffness properties via dual-chamber pressure regulation. Human perception studies with 10 participants reveal that users can distinguish nine discrete states across three size categories and three stiffness levels with 89.4\% accuracy and 6.7~s average response time. We further demonstrate extended architectures that combine AFPAs with complementary pneumatic structures to enable shape or geometry morphing with concurrent stiffness control. Our results establish antagonistic pneumatic principle as a pathway toward next-generation haptic interfaces, capable of multi-dimensiona rendering properties within practical wearable constraints. 
\end{abstract}

\section{Introduction}

Haptic feedback represents a fundamental element in human-robot interaction, enabling users to perceive and manipulate virtual and remote environments through tactile and kinesthetic sensations \cite{Pacchierotti2017GeneralReview}. By conveying physical properties such as stiffness, texture, and temperature, haptic devices bridge the gap between digital information and physical experience~\cite{Kyoung2025FullDOFHapticScience,Fleck2025HapticReview, hou2024tactile, choi2022Temperature, lee2021temperature,see2022TextureReview,heravi2024ModelTexture,qiu2024quantitativeTexture}. This technology has catalyzed breakthroughs across diverse domains, from teleoperation\cite{Patel2022teleoperation,gonzalez2021teleoperation,palagi2023Teleoperation,garavagno2024Teleoperation} and immersive virtual reality systems~\cite{choi2018claw} to robot-assisted rehabilitation therapies~\cite{zbytniewska2019sensorimotorImpairments, ranzani2023Unsupervised-Therapy, devittori2024UnsupervisedRobot-assiste,yeh2017Haptic-Stoke-rehabilitation,zbytniewska2021Sensorimotor-Impairments,Abtahi2018ShapeDisplayMotor} and precision surgical interventions~\cite{Yeon2022ScienceHealtheCare}. 

A critical challenge in haptic rendering lies in simultaneously modulating multiple physical properties (size, shape and stiffness) within a single, portable device. Current haptic interfaces typically excel at rendering either geometric complexity or mechanical properties, but rarely both \cite{Frisoli2024HapticReview}. Mobile robotic systems~\cite{Suzuki2021HapticBotsMovngShape} and shape displays~\cite{Abtahi2018ShapeDisplayMotor, Robertson2018PneumaticSoftShapeStiffness} achieve high-fidelity geometric rendering but require complex mechanical assemblies that preclude wearability. Conversely, portable solutions such as fabric-based devices~\cite{Ota2024Roll-shape} and wearable exoskeletons~\cite{Wang2019KinestheticFeedback, Hinchet2018DextrESClutch, choi2016WolverineGrasping,fontana2013haptichandexoskeleton} prioritize form factor over functional versatility, limiting their capacity for multi-dimensional property modulation.

Soft pneumatic actuators have emerged as a promising paradigm for wearable haptic interfaces, offering inherent compliance, safety, and lightweight construction~\cite{Li2021SoftActuator, Feng2023XPams}. These systems excel at delivering diverse tactile sensations—from guidance cues~\cite{raitor2017WrapGuidance, jumet2023fluidically} to social touch simulation~\cite{yamaguchi2023handshake, salvato2021HumanTouch}—while enabling straightforward stiffness control through pressure modulation~\cite{sebastian2017SoftVariableStiffness, Cosima2024HaptiknitStiffness}. Recent advances have demonstrated shape variation capabilities by arranging multiple pneumatic chambers in specific configurations, allowing switching between actuators with variable stiffness~\cite{jeon2025VaryShape}. Nevertheless, such approaches rely on discrete on/off control schemes, which fundamentally limit the continuity and diversity of achievable shapes. In addition, existing pneumatic haptic devices do not address the simultaneous rendering of multiple physical properties.

Prior attempts at multi-property haptic rendering have yielded partial solutions. Takizawa et al. demonstrated concurrent size and stiffness variation using motorized constraints on pneumatic chambers~\cite{Takizawa2017Size-stiffness}, however, their system required stationary mounting and complex control hardware. Jamming-based approaches enable variable-stiffness shape displays~\cite{stanley2013HapticJammingGeometry, Stanley2016Jamming}, yet suffer from substantial form factors incompatible with wearable applications. Modular pneumatic systems allow object representation through manual reconfiguration~\cite{nguyen2020design, Liu2025PenutouchShape,jeon2025VaryShape}, but cannot dynamically morph between states during interaction.

The recent development of antagonistic fabric-based pneumatic actuators (AFPAs) presents an unexplored opportunity for multi-dimensional haptic rendering. These structures demonstrate remarkable versatility in achieving controllable bending~\cite{Bilodeau2018AntagonisticBending}, contraction~\cite{Jae2024AntagonisticPAMs, Usevitch2018AntagonisticPAMs}, and torsion~\cite{Oh2023AntagonisticTorsional} with integrated stiffness modulation. By leveraging opposing pneumatic chambers, AFPAs can generate complex mechanical behaviors through simple pressure control, suggesting their potential for comprehensive haptic property rendering.

Here, we introduce HapMorph, a novel framework for multi-dimensional haptic rendering based on antagonistic pneumatic architectures. Our approach enables continuous, simultaneous modulation of object size and stiffness within a lightweight, wearable form factor. We demonstrate the approach through a wearable prototype designed as a wearable grasping interface, weighing only 21~g. It utilizes antagonistic pouch motors to deliver continuous size variation from 50 to 104~mm in diameter and stiffness modulation up to 4.7~N/mm through dual-chamber pressure control. Through systematic characterization of the decoupled control of size and stiffness of the interface, we show in a perception study that users can distinguish nine discrete states across three size categories and three stiffness levels with 89.4\% accuracy. Furthermore, we explore extended AFPAs architectures with complementary pneumatic structures to enable shape or geometry morphing with concurrent stiffness control, establishing a pathway toward higher-dimensional haptic interfaces.



\section{Results}

\subsection*{HapMorph System Design and Operation}

\begin{figure}[htbp]
    \centering
    \includegraphics[width=1\textwidth]{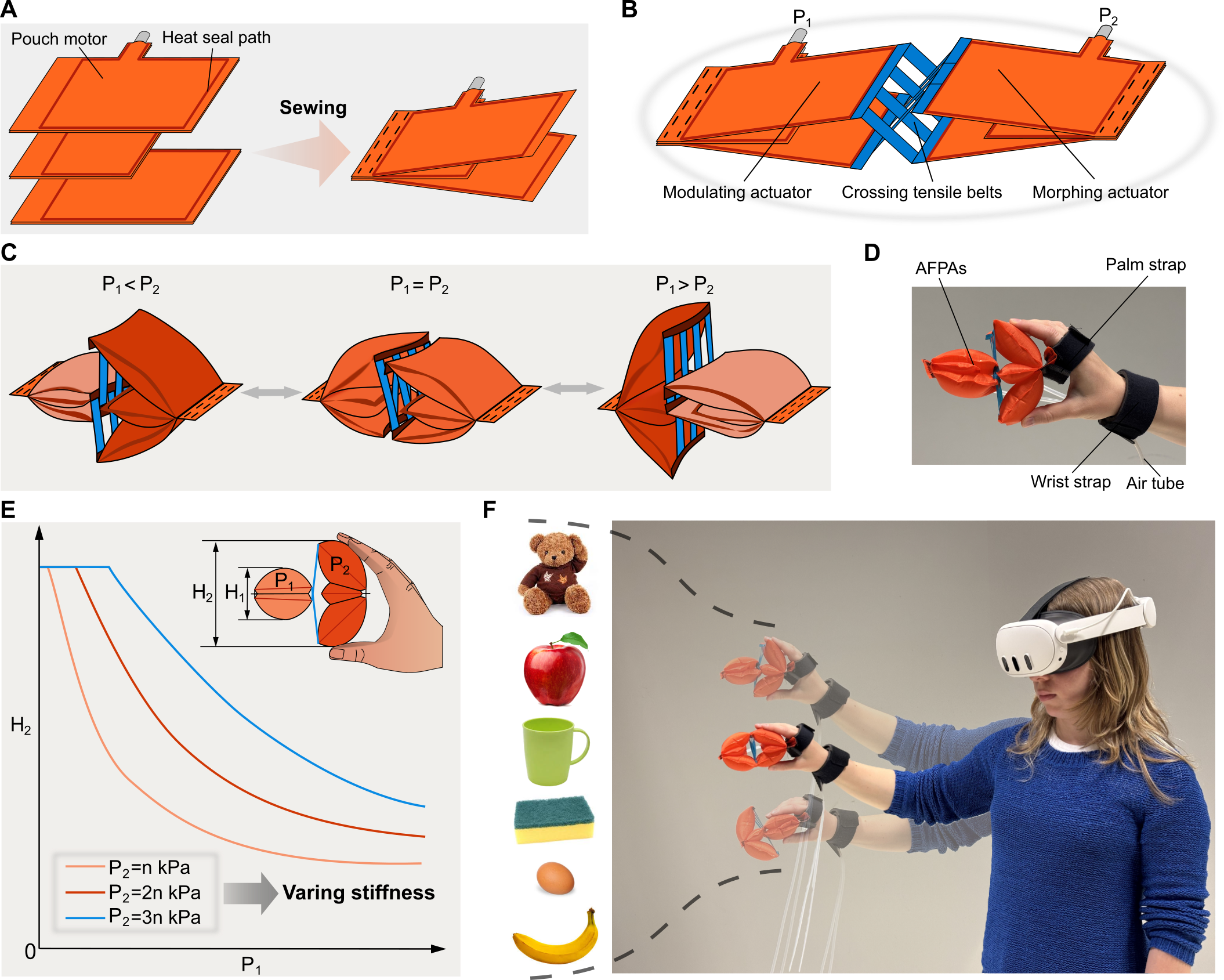}
    \caption{\textbf{HapMorph system overview.}
    \textbf{(A)} Three pouch motors were sewn to form a "one-sided" expansion actuator. \textbf{(B)} The AFPAs comprise two antagonistically arranged sewn actuators with crossing belts, independently actuated via control pressures $P_1$ and $P_2$.
    \textbf{(C)} By tuning the internal pressures $P_1$ and $P_2$, the AFPAs exploit the mechanical interplay between the two actuator groups to dynamically modulate overall shape and stiffness. \textbf{(D)} The components of the HapMorph device. \textbf{(E)} Modulation of $P_2$—designated as the haptic output side—primarily governs the output stiffness perceived by the user, while variation in $P_1$ enables continuous adjustment of the device height on the $P_2$ side. \textbf{(F)} When integrated into virtual reality (VR) applications, the HapMorph platform enables users to perceive and differentiate among virtual objects of varying dimensions and stiffness. 
    }
    \label{Fig-M-Illustration}
\end{figure}

To enable multi-dimensional haptic rendering, we implemented the AFPAs using thermoplastic polyurethane (TPU) coated fabric. We fabricated three pouch motors (Fig. \ref{Fig-M-Illustration}A and Fig. S1A) \cite{niiyama2014pouch,niiyama2015pouch}. The middle layer pouch motor was designed with a shorter length to prevent interference with the tensile belts during operation (Fig. S1B). 
We sewed one side of the three pouch motors together to create a "one-sided" expansion actuator that maintains structural integrity while allowing controlled deformation (Fig. \ref{Fig-M-Illustration}A and Fig. S1C). 

Two expansion actuators were connected by crossing tensile belts to create AFPAs (Fig. \ref{Fig-M-Illustration}B, Fig. S1D and E). The right expansion actuator serves as the morphing actuator, representing the physical interface for the user, while the left expansion actuator works as the modulating actuator, controlling the behavior of the morphing actuator. The introduction of the antagonistic configuration enables simultaneous control of multiple haptic properties. When both actuators are inflated, the tensile belts limit the expansion of both actuators until force equilibrium is achieved, creating a mechanically coupled system where the interaction between opposing forces determines the final configuration.

By regulating the pressure of the two actuators ($P_1$ for modulating actuator and $P_2$ for morphing actuator), both the height and stiffness of the morphing actuator can be rendered continuously and independently (Fig. \ref{Fig-M-Illustration}C and E). 

The AFPAs interface is worn at the user's hand through a hand-palm strap and a wrist strap, as shown in Fig. \ref{Fig-M-Illustration}D. Two soft silicone air tubes are routed through the wrist strap to a remote pneumatic driving unit.

As shown in Fig. \ref{Fig-M-Illustration}F, the designed HapMorph interface addresses fields such as virtual reality and teleoperation rendering, enabling realistic simulation of the stiffness and size properties of the interacting objects, with minimal user fatigue due to the lightness of the wearable parts.

\subsection*{Height morphing characterization}

\begin{figure}[htbp]
    \centering
    \includegraphics[width=1\textwidth]{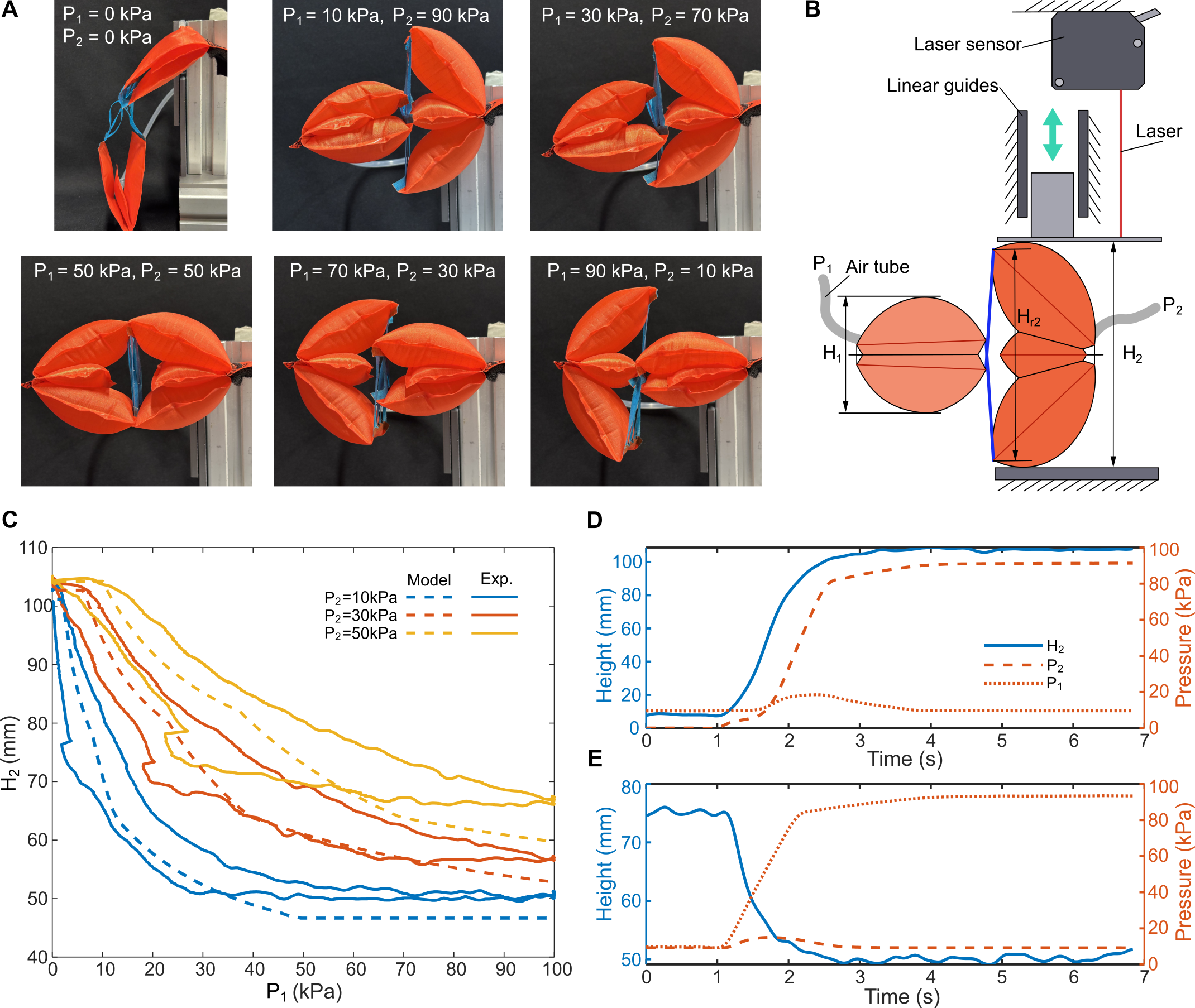}
    \caption{\textbf{Height morphing characterization.} \textbf{(A)} By independently regulating the input pressures $P_1$ and $P_2$, the HapMorph device modulated its size across a substantial range. \textbf{(B)} Schematic of the experimental setup for characterizing the relationship between actuation pressure ($P_1$ and $P_2$) and height ($H_2$). \textbf{(C)} Both experimental data and model predictions were compared to evaluate the influence of $P_1$ modulation under constant $P_2$ conditions. \textbf{(D)} and \textbf{(E)} present the step response characteristics of the HapMorph system to pressure inputs $P_1$ and $P_2$, respectively, showing rapid response times.}
    \label{Fig-M-Size}
\end{figure}

We characterized the relationship between height ($H_2$ for finger interaction) and pressures ($P_1$ for the modulating actuator and $P_2$ for the morphing actuator) through comprehensive experimental analysis, as shown in Fig. \ref{Fig-M-Size}. Under different pressure combinations for $P_1$ and $P_2$, the AFPAs exhibited distinctly different heights, demonstrating the system's ability to achieve continuous size modulation. Compared to the opposing actuator, higher pressure consistently resulted in greater height expansion, following predictable mechanical principles (Fig. \ref{Fig-M-Size}A). 

To predict and understand HapMorph's behavior, we developed a mathematical model based on the virtual work principle (see Supplementary Text and Fig. S2)\cite{Feng2023XPams,niiyama2014pouch, niiyama2015pouch, yang2024highPAMs}. We then validated this theoretical framework using the experimental setup shown in Fig. \ref{Fig-M-Size}B and Fig. S3A, which enables precise measurement of actuator height under controlled pressure conditions.

The experimental results, presented in Fig. \ref{Fig-M-Size}C, demonstrate excellent agreement between model predictions and measured data. The model accurately predicted HapMorph behavior across the full range of morphing pressure $P_2$ values. For a given pressure $P_2$ (morphing chamber), the height of the morphing actuator ($H_2$) decreased monotonically with increasing pressure of the modulating actuator ($P_1$), reflecting the antagonistic nature of the system. Conversely, for a given pressure $P_1$, height $H_2$ increased predictably with increasing $P_2$. The achievable size range spans from 50~mm to 104~mm, providing substantial coverage for most everyday objects encountered in daily interactions. The maximum height $H_2$ is slightly higher with higher morphing pressure $P_2$, which is attributed to the elastic deformation of the tensile belts under higher pressure.
Hysteresis between loading and unloading cycles was observed due to residual friction between chambers and possible plastic deformation of the fabric materials. The hysteresis slightly increased with increasing morphing pressure ($P_2$). 

We systematically measured the pressure step response of the HapMorph device to characterize its dynamic performance. As shown in Fig. \ref{Fig-M-Size}D, with pressure $P_1$ maintained constant while pressure $P_2$ was instantly increased from 0 to 90~kPa, the system demonstrated rapid response characteristics. After approximately 2~seconds (limited primarily by air supply speed\cite{joshi2021pneumaticSupply} rather than actuator dynamics), the height increased from 10~mm to 103~mm, representing substantial size change. Conversely, with fixed pressure $P_2$ (10~kPa), height $H_2$ decreased rapidly from approximately 75~mm to 50~mm in approximately 2~seconds when pressure $P_1$ increased from 10~kPa to 90~kPa (Fig. \ref{Fig-M-Size}E). The results demonstrate that actuator height modulation can be achieved within seconds, typically within 2~seconds, making the system suitable for real-time interactive applications.

\subsection*{Stiffness morphing characterization}

\begin{figure}[htbp]
    \centering
    \includegraphics[width=0.9\textwidth]{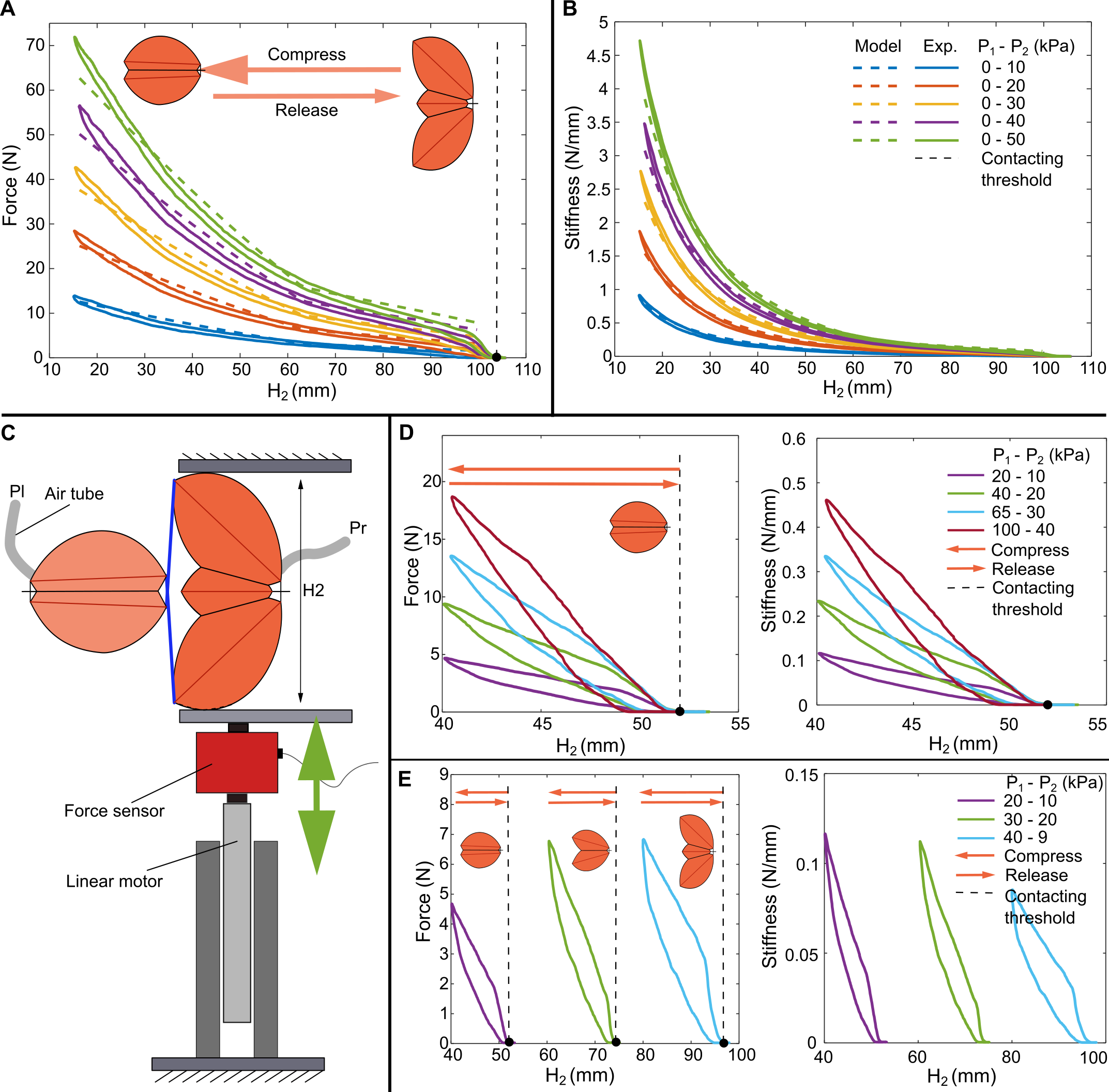}
    \caption{\textbf{Stiffness morphing characterization.} \textbf{(A)} and \textbf{(B)} present both simulated and experimental results characterizing the relationship between force versus deformation and stiffness versus displacement under varying $P_2$ pressure conditions. \textbf{(C)} Schematic of the experimental setup used to measure stiffness variations under different pressure levels. \textbf{(D)} By configuring distinct combinations of $P_1$ and $P_2$, the HapMorph device can achieve a range of stiffness values even at a fixed output size, enabling independent control of size and compliance. \textbf{(E)} Furthermore, appropriate tuning of $P_1$ and $P_2$ allows the system to exhibit similar stiffness levels across varying output sizes, demonstrating the decoupling capability between size and mechanical compliance.}
    \label{Fig-M-Stiffness}
\end{figure}

We characterized the effects of pressures ($P_1$ and $P_2$) on the stiffness of the morphing actuator using a comprehensive experimental approach with the setup shown in Fig. \ref{Fig-M-Stiffness}C and Fig. S4A. The experimental protocol involved the setting of the pressure of the modulating actuator ($P_1$) to 0 and the pressure of the morphing actuator ($P_2$) to various constant values. We then applied controlled compression by slowly reducing the height $H_2$ of the morphing actuator of 15~mm before returning to the starting point, while continuously monitoring the force response. 

The experimental results were systematically compared with our theoretical modeling predictions (see Supplementary Text) and demonstrated excellent alignment across the tested parameter space (Fig. \ref{Fig-M-Stiffness}A and B). Both force and stiffness increased nonlinearly with decreasing actuator height ($H_2$), exhibiting the characteristic behavior of pneumatic actuators where smaller volumes result in higher effective stiffness (up to approximately 4.7~N/mm when $H_2$ = 15~mm). The stiffness curve is relatively flat when $H_2$ is large, and the stiffness curve becomes steeper as $H_2$ decreases. This behavior occurs because as $H_2$ decreases, the contact area between pouch motors increases, resulting in higher stiffness. The stiffness variation depends on both the pressure of the morphing actuator ($P_2$) and the instantaneous morphing actuator height, creating a rich parameter space for haptic control (Fig. \ref{Fig-M-Stiffness}B).

We systematically characterized the ability to independently control stiffness and position for HapMorph through extensive parameter mapping (Fig. S4B and S4C). By appropriately selecting four representative pressure combinations for the modulating actuator ($P_1$) and morphing actuator ($P_2$), HapMorph can present perceptable different stiffness values ranging from 0.12~N/mm to 0.46~N/mm, spanning nearly a 4-fold range (Fig. \ref{Fig-M-Stiffness}D). 

Crucially, with appropriate pressure combinations, it is possible to morph the size of the morphing actuator while maintaining approximately constant stiffness (Fig. \ref{Fig-M-Stiffness}E). At three different actuator heights (52~mm, 73~mm, and 96~mm), the actuator exhibited remarkably similar stiffness levels (0.1~N/mm ± 0.015~N/mm), demonstrating the system's potential for truly independent control of size and mechanical properties. This decoupling capability represents a significant advancement over existing haptic devices that typically exhibit coupled size-stiffness relationships. Additional pressure combinations for $P_1$ and $P_2$ were systematically investigated to establish comprehensive stiffness-pressure relationships, providing a complete characterization of the achievable haptic property space (Fig. S4B and C).

\subsection*{Extended architectures for enhanced functionality}

\begin{figure}[htbp]
    \centering
    \includegraphics[width=0.9\textwidth]{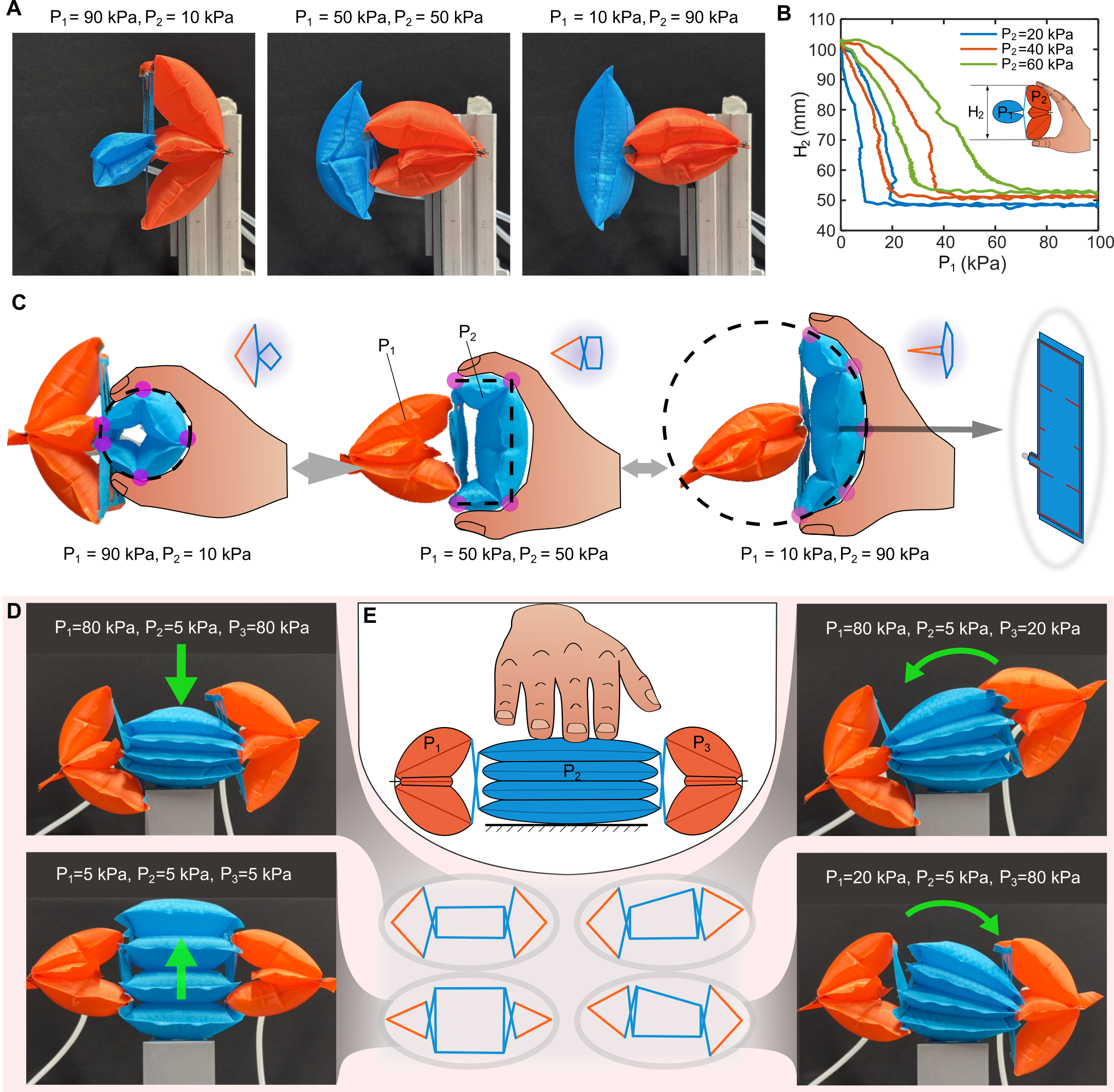}
    \caption{\textbf{Extended architectures for enhanced functionality.} \textbf{(A)} HapMorph compatibility with alternative pneumatic actuator configurations. \textbf{(B)} Height variation under differential pressure conditions ($P_1$ and $P_2$), showing enhanced linearity with increased hysteresis. \textbf{(C)} Redesigned interaction interface modulates deformation sequencing, enabling diverse shape and stiffness feedback including circular and rectangular morphologies. \textbf{(D)} Multiple pouch motor assemblies provide enhanced DOF for haptic delivery, conveying geometrical cues through coordinated control. \textbf{(E)} Multi-DOF feedback demonstration with complex directional information through finger perception mechanisms.}
\label{Fig-M-Extended}
\end{figure}

Beyond the basic dual-actuator symmetrical configuration, alternative forms of pneumatic expansion actuators can effectively modulate size and stiffness. With the parameters shown in Fig. S5, we implemented a single pouch motor \cite{Feng2023XPams} configuration as a modulating actuator to control the size and stiffness of the morphing actuator. The morphing height ($H_2$) can be precisely controlled by adjusting both the modulating pressure ($P_1$) and morphing pressure ($P_2$), as demonstrated in Fig. \ref{Fig-M-Extended}A and B. Despite achieving a more linear relationship between $P_2$ and $H_2$ during height changes, this configuration exhibits significant hysteresis between loading and unloading cycles, presenting additional challenges for precise height control that require advanced control algorithms to compensate.

Extending beyond simple size control, we can continuously morph both shape and stiffness simultaneously through strategic actuator design. The morphing actuator incorporates internal constraint lines with carefully designed lengths to control cross-sectional deformation patterns (Fig. S6). By optimizing the geometry of these internal constraints, the cross-sectional area can be dynamically controlled during operation. When the modulating actuator is pressurized, the morphing actuator bends sequentially around the constraint lines in a predetermined sequence that depends on the designed cross-sectional area distribution (Fig. \ref{Fig-M-Extended}C). By decreasing the modulating actuator pressure ($P_1$), the morphing actuator undergoes controlled shape transformation from a small cylinder to a cuboid configuration, and finally to a large cylinder, providing users with distinct shape feedback modalities.

To achieve higher degrees of freedom in haptic rendering, we explored novel configurations with additional modulating actuators and tensile belt networks (Fig. \ref{Fig-M-Extended}D). Two independent groups of pouch motors were employed as modulating actuators, mechanically connected to a 4-layer morphing actuator through a carefully designed tensile belt system (Fig. S7). With two modulating actuators and one morphing actuator, the device can dynamically display vertical and lateral displacement, while maintaining variable stiffness control. This multi-dof capability enables simulation of more complex object interactions and directional force feedback. One envisaged potential application involves the rendering of surface orientation in addition to displacement and stiffness (Fig. \ref{Fig-M-Extended}E). 

\subsection*{Human perception study}

\begin{figure}[htbp]
    \centering
    \includegraphics[width=1\textwidth]{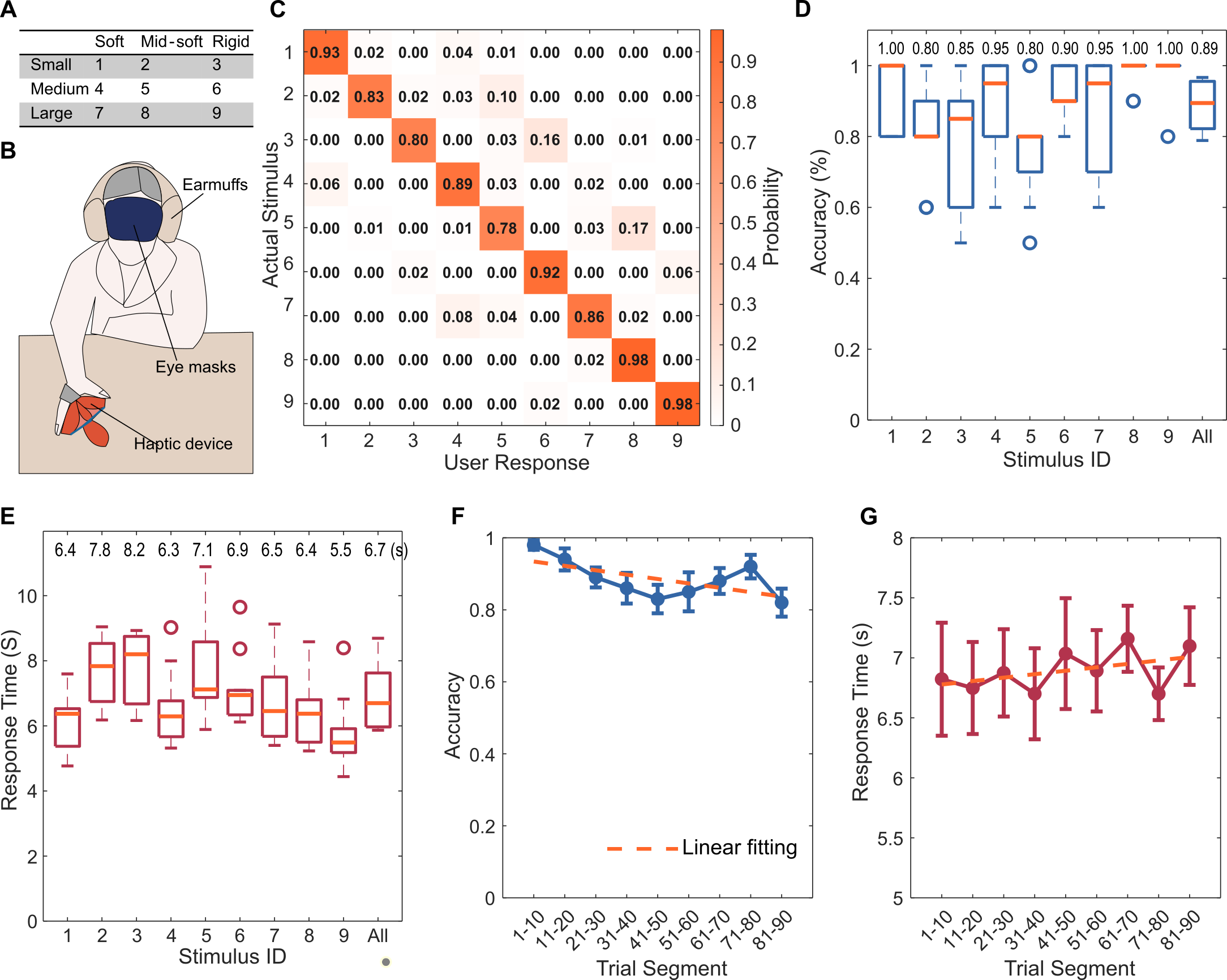}
    \caption{\textbf{Human perception study.} \textbf{(A)} Correspondence between states 1-9 and their respective dimensional and stiffness parameters, spanning three size categories (small, medium, large) and three stiffness levels (soft, medium, hard). \textbf{(B)} Experimental configuration for the user study: participants were asked to identify nine distinct states of HapMorph relying solely on haptic perception. \textbf{(C)} Confusion matrix depicting the classification results from the perceptual discrimination task, showing systematic patterns in identification accuracy. \textbf{(D)} and \textbf{(E)} Distribution of identification accuracy and response latency across the different haptic states, revealing individual differences and state-dependent performance variations. \textbf{(F)} and \textbf{(G)} Performance metrics across trial segments, showing systematic changes in accuracy and response time with trial progression.}
    \label{Fig-M-Subjects}
\end{figure}

We conducted a comprehensive human perception study with 10 healthy participants to evaluate HapMorph's ability to convey distinguishable haptic information across multiple property dimensions. Participants were equipped with opaque blindfolds and noise-canceling headphones to eliminate external visual and auditory cues (Fig. \ref{Fig-M-Subjects}B). Participants were asked to distinguishing between nine different states of the haptic interface relying solely on their grasping perception. The nine states were a combination of three sizes (small, medium, and large) and three stiffness levels (soft, medium, and hard) (Fig. \ref{Fig-M-Subjects}A). Each subject was asked to complete 90 trials (10 repetitions for each state in random sequence) in a single session. 

The confusion matrix of all test results is illustrated in Fig. \ref{Fig-M-Subjects}C, revealing the system's effectiveness under purely haptic perception conditions. Different states exhibited varying accuracy rates, with performance ranging from 78\% to 98\% across the nine tested conditions. The lowest accuracy occurred for the medium size and medium stiffness combination (state 5) at 78\%, which represents the most challenging discrimination task where both properties fall in intermediate ranges. In contrast, extreme combinations achieved the highest accuracy, reaching 98\% for states with clearly differentiated size and stiffness characteristics. Detailed analysis of the confusion matrix revealed that participants made systematically fewer errors in size discrimination compared to stiffness discrimination, with the most of the incorrect responses occurring between different stiffness levels rather than between different sizes. 

Fig. \ref{Fig-M-Subjects}D compares the accuracy distribution across different states among users, demonstrating considerable individual variation in perceptual capabilities. The overall accuracy for all the states was 89.4\%, demonstrating the effectiveness of HapMorph in delivering distinguishable size and stiffness information across the tested parameter space. Perception of size and stiffness varied systematically between users, particularly for intermediate states 3 and 7, suggesting individual differences in haptic sensitivity and exploration strategies. For certain clearly differentiated states such as 1, 8, and 9, most participants achieved 100\% accuracy, with only occasional outliers experiencing difficulty.

Fig. \ref{Fig-M-Subjects}E presents the temporal characteristics of state identification, comparing the time required for subjects to identify different haptic states. On average, the response time was 6.7~s for all subjects, which includes multiple components: the duration needed for haptic exploration and state discrimination, the mechanical transition time for HapMorph to change states (approximately 2~s according to Fig. \ref{Fig-M-Size}D and E), and the administrative time for the recorder to signal, receive feedback, and document results. State identification predominantly required 5-9~seconds, with systematic variation depending on state complexity. State 9 had the shortest identification time (median $\approx 5.5$~s, standard deviation $\approx 0.8$~s), whereas State 3 required the longest (median $\approx 8.3$~s, standard deviation $\approx 2$~s), reflecting the increased difficulty in distinguishing intermediate property combinations.

To assess learning effects and fatigue, we segmented the 90 trials into 9 sequential segments (10 trials per segment) and calculated average performance metrics for each segment. By comparing performance across trial segments, we evaluated the effects of time and experience on accuracy and response time. As shown in Fig. \ref{Fig-M-Subjects}F and G, accuracy decreased systematically from nearly 1.0 to 0.8 with increasing trial number, while response time increased gradually with trial progression. Such performance drop in the first half of the experiment might be due to the fact participants were exposed to a growing variety of states (randomized presentation order) and possibly the could have gradually lost confidence in their discrimination criteria, hesitating more in their responses, resulting in both lower accuracy and longer response times. 

Overall, the perception study assessed the ability to perceive size and stiffness cues from HapMorph with high accuracy 89.4\% within practical time constraints (6.7~s average response time), establishing HapMorph's significant potential for conveying diverse haptic information in real-world applications. 

\section{Materials and Methods}

\subsection*{Materials and Fabrication of HapMorph Device}

Thermoplastic polyurethane (TPU) coated fabric (Adventure Expert) was used throughout this work, with 40D ripstop nylon TPU fabric employed for components depicted in orange colour, and 70D ripstop nylon TPU fabric for components depicted in blue. 

A computer numerical control (CNC) thermal sealing system, adapted from a commercial 3D printer platform (Anycubic Mega), was used to fabricate the pouch motors. Air tubes were secured to each chamber using cyanoacrylate adhesive (Fig. S1A). In parallel with chamber fabrication, the 70D TPU-coated fabric was laser-cut into two tensile belt segments and crossed the tensile belts for subsequent assembly steps (Fig. S1B).

The three pouch actuators were sewn on one side, as shown in Fig. S1C. We then arranged two sewn pouch actuator units in opposing orientations to create an antagonistic configuration. These actuators were thermally bonded to the crossed tensile belt system through heat sealing (Fig. S1D and E). Finally, the AFPAs were sewn to elastic fabric, and Velcro was used to fasten the AFPAs to the hand.

\subsection*{Pneumatic Sources and Pressure control}

In all experiments, pneumatic pressure was supplied by a commercial air compressor (FIAC F6000/50) capable of delivering consistent pressure up to 800~kPa with minimal fluctuation. Proportional solenoid valves ITV0010 (SMC) were used to regulate actuator pressure. An ESP32 microcontroller (Espressif Systems) was programmed to provide real-time pressure control with feedback from integrated pressure sensors, ensuring stable and responsive operation throughout all experimental protocols.

\subsection*{Size-pressure Relationship Characterization}

As shown in Fig. \ref{Fig-M-Size}B and Fig. S3A, the actuator was positioned between a fixed base and a lightweight slider (19~g mass) to enable controlled height measurements. A SICK OD2000 laser displacement sensor was employed to continuously monitor actuator height throughout pressure variations. To prevent slipping between the actuator and mechanical components, double-sided adhesive tape was placed on both the slider and base surfaces, above and below the actuator contact areas. 

During the experiments, the morphing actuator pressure ($P_1$) is maintained at a constant value, while the modulating actuator pressure ($P_2$) is decreased from 100~kPa to 0 and subsequently increased back to 100~kPa. Data from the pressure control valves and the displacement laser sensor was synchronously recorded at 16~Hz sampling frequency to capture the dynamic behavior during pressure transitions.

\subsection*{Force-pressure Relationship Characterization}

To systematically apply controlled forces to the pneumatic actuator, a T16 linear actuator (Actuonix Motion Devices) equipped with an internal potentiometer for precise displacement measurement was employed (see Fig. \ref{Fig-M-Stiffness}C and Fig. S4A). Force measurements were obtained using a calibrated LSB205 force sensor (Futek), positioned between the linear actuator and the pneumatic actuator to capture the complete force-displacement relationship. 

During the test, both pressures $P_1$ and $P_2$ were maintained at constant values. The linear actuator was programmed to move upward from an initial position to compress the morphing actuator, returning to its starting position after reaching a specified displacement, with continuous recording of displacement and force data. All mechanical testing was conducted at controlled displacement rates (1.5~mm/s) to ensure quasi-static conditions and eliminate dynamic effects from the stiffness characterization.

\subsection*{User Study Protocol}

Based on comprehensive preliminary investigations of pressure parameter combinations (Fig. S4), we systematically defined 9 distinct haptic states (numbered 1 to 9) for the HapMorph evaluation, each corresponding to specific pressure combinations detailed in Fig. \ref{Fig-M-Subjects}A and Fig. S7A. These states were designed to represent three size categories (small, medium, and large) combined with three stiffness levels (soft, medium, and hard), simulating the range of mechanical properties found in common everyday objects.

We conducted a controlled user study with 10 healthy participants (6 males and 4 females, age 29.2 ± 4.18 years) to evaluate the system's ability to convey distinguishable haptic information. All participants provided written informed consent prior to participation, and the study protocol was approved by the ethical review board of the Scuola Superiore Sant'Anna. Participants were equipped with opaque blindfolds and noise-canceling headphones to eliminate visual and auditory cues, while maintaining HapMorph contact with their dominant right hand (all subjects reported being right-handed). Before the formal experiment began, participants were given some time to familiarize different states.

The experimental protocol required subjects to distinguish the correct haptic state relying solely on tactile, kinesthetic and proprioceptive perception of the hand, as illustrated in Fig. \ref{Fig-M-Subjects}B. A trained experimenter managed state transitions in randomized order and documented subject responses to ensure consistent experimental conditions (Fig. S7B and C). Upon receiving an audio signal, participants started exploration of the presented haptic state for up to 30~seconds before verbally reporting their assessment to the recorder. Following a standardized 5-10 minute familiarization period to reduce learning effects, each subject completed 90 experimental trials (10 repetitions per state) in a single session, to minimize between-session variability while avoiding excessive fatigue. 

Box plots show the median (orange center line), interquartile range (box), and 1.5$\times$IQR whiskers, with outliers as individual points. Independent t-tests were applied to analyze the influence of gender on accuracy and response time, with $p < 0.05$ considered statistically significant.

\section{Discussion}

In this work we proposed and explored a AFPAs based approach to develop lightweight and multi-modal haptic interfaces. 
Considering the fabrication process, we implemented a working prototype  using affordable materials and fabrication technologies \cite{Feng2023XPams}. These consisted in TPU-coated and elastic fabrics, CNC heat sealing and lasercut manufacturing. The HapMorph prototype resulted in a lightweight (21~g) wearable device that can render combined multi-dimensional physical properties (size and stiffness modulation).

In our characterization studies, we demonstrated that the behavior of HapMorph can be accurately predicted by the developed mathematical model. The size (height of the morphing actuator) can be controlled by regulating the pressures for the modulating actuator and morphing actuator, although some limited hysteresis is observed in the operation cycles.

In the conducted stiffness characterization, we measured the stiffness of the morphing actuator under various conditions. With the morphing actuator height of 15~mm, the stiffness could reach up to 4.7~N/mm (potentially higher with increased morphing pressure), which is comparable to previously reported research\cite{Takizawa2017Size-stiffness}. The maximum force recorded was 72~N at this height, which is relatively high in comparable studies. However, the stiffness depends on both the pressure and height of the morphing actuator. At greater heights, the stiffness output approaches 0.1~N/mm due to limited contact area between actuators, Although this performance could be improved by implementing larger actuators or higher morphing pressure.

The extended architectures provide additional versatility for HapMorph implementation. The system can work with other types of expansion actuators connected via tensile belts, or high-stroke pneumatic artificial muscles for enhanced modulating range. Shape morphing with concurrent stiffness control is achievable by implementing constraint lines (curves or specific geometries) to control the actuator's volume change, resulting in sequential bending patterns. In this work, we used only 3 constraint lines to segment the chamber; additional lines would enable more complex shape delivery. Multiple actuators provide enhanced degrees of freedom, enabling more sophisticated morphing capabilities such as complex geometry rendering. I.e. we demonstrated that two modulating actuators with one morphing actuator could deliver multi-dimensional geometric information (height and lateral orientation). 

Our human perception study reveals important insights into multi-dimensional haptic perception. The 89.4\% overall accuracy in distinguishing nine discrete states demonstrates that users can effectively perceive and discern combinations of size and stiffness properties. The overall response time of 6.7~s includes the time for HapMorph morphing transients, subjects' exploration and decision-making, as well as administrative time for recording results and switching states. Accounting for the contribution of recording and HapMorph morphing (approximately 2~seconds), the actual perceptual response time can be estimated at 2-4~seconds depending on stimulus complexity. Due to fatigue effects and the absence of feedback, overall performance of subjects decreased with trial progression, resulting in lower accuracy and slightly increased response times.

We also collected anthropometric data including subjects' hand length (defined as the distance from the tip of the middle finger to the midline of the distal wrist crease when the forearm and hand are supinated on a table)\cite{hager2000handlength, fallahi2011handlength} to investigate potential correlations with performance. The effects of hand length and gender on both accuracy and response time were systematically investigated through statistical analysis (please see Fig. S8, Fig. S9, and Fig. S10). However, no significant correlations were found between anthropometric factors and performance metrics, suggesting that HapMorph's effectiveness is robust across diverse user populations.

Several limitations warrant consideration for future development. The current response time of approximately 2~seconds, while it is acceptable for many applications, could be improved through optimized pneumatic supply systems and reduced actuator volumes. Regarding the limited observed hysteresis, it presents challenges if a more precise control is desirable for certain applications, and may require advanced compensation algorithms. The extended architectures require systematic characterization as well as comprehensive user studies. Regarding device fabrication, future lifecycle testing of HapMorph components could identify structural weaknesses and guide optimization efforts. Additionally, the system's current reliance on external pneumatic supply limits true portability, suggesting opportunities for developing integrated pneumatic generation systems. Future work should focus on integrating micro-pumps, lightweight valves, and energy-efficient control algorithms to achieve fully autonomous, battery-powered operation for mobile applications.

The demonstrated capability for independent size and stiffness control, combined with lightness of the wearable parts, represents a significant advancement in haptic interface design. Unlike previous single-chamber pneumatic systems that exhibit inherent coupling between geometric and mechanical properties, HapMorph achieves true decoupling through its antagonistic pneumatic architecture. This capability opens new possibilities for haptic rendering in applications requiring simultaneous modulation of multiple object properties, such as virtual reality training systems, teleoperation interfaces, and assistive technologies.

Future research directions include extending the framework to additional haptic modalities such as texture and temperature, developing closed-loop control algorithms for improved precision, and exploring miniaturization strategies for enhanced portability. The integration of sensing capabilities could enable adaptive haptic rendering based on user interactions and preferences.

In conclusion, our work demonstrates that antagonistic pneumatic actuator principles can overcome fundamental limitations in haptic device design, enabling simultaneous modulation of multiple object properties within practical wearable constraints. The HapMorph framework represents a significant advancement in human-machine interaction by achieving independent control of size and stiffness through a mechanically coupled yet controllably decoupled system. This technology establishes a foundation for next-generation haptic interfaces capable of conveying the rich complexity of physical interactions in virtual and augmented environments.



\section*{Acknowledgments}
We thank the participants in our user study for their time and dedication. We also acknowledge valuable discussions with colleagues in the field of haptic interfaces and soft robotics that helped shape this research.


\subsection*{Author Contributions} 

Conceptualization: R. Chen, D. Leonardis, D. Chiaradia, A. Frisoli.

Methodology and Investigation: R. Chen, D. Leonardis.

Writing—original draft: R. Chen.

Writing—review \& editing: R. Chen, D. Leonardis, D. Chiaradia, A. Frisoli.

Funding acquisition and administration: AF, D. Leonardis.

\subsection*{Funding}

Daniele Leonardis is funded by the SUN project, supported by the European Union's Horizon Europe research and innovation program under Grant Agreement No. 101092612. Domenico Chiaradia is funded by the Next Generation EU project ECS00000017 "Ecosistema dell'Innovazione" Tuscany Health Ecosystem (PNRR, Spoke 4: Spoke 9: Robotics and Automation for Health). Rui Chen and Antonio Frisoli are partially supported by MSCA-DN / Project 101073374 – ReWIRE, and partialy the same Next Generation EU project ECS00000017.

\subsection*{Conflicts of Interest}
The authors declare that there is no conflict of interest regarding the publication of this article.

\subsection*{Data Availability}

All data are available in the main text or the supplementary materials. The fabrication protocols, control software, and experimental datasets supporting the conclusions of this article are available from the corresponding author upon reasonable request.

\section*{Supplementary Materials}
Supplementary Text\\
Figs. S1 to S10\\
Movies S1 to S5 

\printbibliography

\end{document}